\documentclass[10pt,twocolumn,letterpaper]{article}

\usepackage{iccv}
\usepackage{times}
\usepackage{epsfig}
\usepackage{graphicx}
\usepackage{amsmath}
\usepackage{amssymb}

\usepackage{enumitem}
\usepackage{array}
\usepackage{float}
\usepackage{algorithm}
\usepackage{algorithmic}
\usepackage{color}

\newcolumntype{C}[1]{>{\centering\arraybackslash}p{#1}}
\floatname{algorithm}{Algorithm}

\graphicspath{ {Figures/} } 

\usepackage[pagebackref=true,breaklinks=true,letterpaper=true,colorlinks,bookmarks=false]{hyperref}

\iccvfinalcopy 

\def\iccvPaperID{626} 

\ificcvfinal\fi

\begin{document}

\title{Infinite Latent Feature Selection:\\A Probabilistic Latent Graph-Based Ranking Approach}

\author{Giorgio Roffo\\
University of Glasgow\\
{\tt\small Giorgio.Roffo@Glasgow.ac.uk}
\and
Simone Melzi\\
University of Verona\\
{\tt\small Simone.Melzi@univr.it}
\and
Umberto Castellani\\
University of Verona\\
{\tt\small Umberto.Castellani@univr.it}
\and
Alessandro Vinciarelli \\
University of Glasgow\\
{\tt\small Alessandro.Vinciarelli@Glasgow.ac.uk}\\
}
       
\maketitle

\begin{abstract}
Feature selection is playing an increasingly significant role with respect to many computer vision applications spanning from object recognition to visual object tracking. However, most of the recent solutions in feature selection are not robust across different and heterogeneous set of data. In this paper, we address this issue proposing a robust probabilistic latent graph-based feature selection algorithm that performs the ranking step while considering all the possible subsets of features, as paths on a graph, bypassing the combinatorial problem analytically. An appealing characteristic of the approach is that it aims to discover an  abstraction behind low-level sensory data, that is, relevancy. Relevancy is modelled as a latent variable in a PLSA-inspired generative process that allows the investigation of the importance of a feature when injected into an arbitrary set of cues. The proposed method has been tested on ten diverse benchmarks, and compared against eleven state of the art feature selection methods. Results show that the proposed approach attains the highest performance levels across many different scenarios and difficulties, thereby confirming its strong robustness while setting a new state of the art in feature selection domain.  
\end{abstract}


%
%
\section{Introduction}\vspace{-0.2cm}

Performance of machine learning methods is heavily dependent on the choice of features on which they are applied. Different features can entangle and hide the different explanatory factors of variation behind the data. Feature Selection (FS) aims at improving the performance of a prediction system, allowing faster and more cost-effective models, while providing a better understanding of the inherent regularities in data. In the recent \emph{computer vision} literature there are many scenarios where FS is a crucial operation \cite{FS_ICCV1,Roffo:InfFS:2015,FS_ICCV3,FS_ICCV2,ICCV_4,RoffoBMVC2016}. From multiview face recognition \cite{FS_ICCV2} where FS is used to speed up the multiview face recognition process and to maintain the generalization performance, to object recognition \cite{Roffo:InfFS:2015}, until real-time visual object tracking \cite{RoffoBMVC2016,KristanLMFPCVHL16} where FS dynamically identifies discriminative features that help in handling the appearance variability of the target by improving tracking performance.

In this paper, we propose a probabilistic latent graph-based feature selection algorithm that performs the ranking step by considering all the possible subsets of features exploiting the convergence properties of power series of matrices. We map the feature selection problem to an affinity graph (e.g., feature $\approx$ node), and then we consider a subset of features as a path connecting set of nodes. An appealing characteristic of the approach is that the importance of a given feature is modelled as a conditional probability of a latent variable and features, namely $P(z|f)$. Our approach aims to model an important hidden variable behind data, that is, \emph{relevancy} in features. Raw values are observable while relevancy to a particular task is not (e.g., in classification), therefore, relevancy is modelled as an abstract latent variable. In particular, our approach consists of three main parts:\vspace{-0.2cm}
\begin{itemize}
    \item \textbf{Pre-processing}: a quantization process is applied on raw feature  distributions $\vec x_i$, mapping their values to a countable nominal smaller set of tokens. The pre-processing step assigns a descriptor $f_i$ to each raw feature $\vec x_i$.\vspace{-0.2cm}
    \item \textbf{Graph-Weighting}: we build an undirected fully-connected graph, where nodes correspond, one by one, to each feature $f_i$, and each weighted edge among $f_i \leadsto f_j$ models the probability that features $x_i$ and $x_j$ are relevant. Weights are learnt automatically by a learning framework based on a variation of the probabilistic latent semantic analysis (PLSA) technique~\cite{hofmann1999probabilistic}, which models the probability of each co-occurrence in $f_i, f_j$ as a mixture of conditionally independent multinomial distributions. Parameters are estimated using the Expectation Maximization (EM) algorithm.  \vspace{-0.2cm}
    \item \textbf{Ranking}: the ranking step is done following the idea of the Infinite Feature Selection (Inf-FS) \cite{Roffo:InfFS:2015}, that considers all the possible paths among nodes investigating the redundancy of any features when injected into arbitrary sets of cues.\vspace{-0.2cm}
\end{itemize}

The proposed method is compared against 11 state of the art feature selection methods selected from recent literature in the machine learning and pattern recognition domains, reporting results for a total of $576$ unique tests (note, the source code is available at \href{https://goo.gl/uTuZhc}{Matlab-Central}). We selected 10 publicly available benchmarks of cancer classification and prediction on DNA microarray data (\emph{Colon}~\cite{alon}, \emph{Lymphoma}~\cite{Golub99}, \emph{Leukemia}~\cite{Golub99}, \emph{Lung}~\cite{Gordon02},  Prostate~\cite{citeulike:1624492}), handwritten character recognition (GINA~\cite{GINA}), text classification from the NIPS feature selection challenge (DEXTER ~\cite{NIPS2003}), and a movie reviews corpus for sentiment analysis (\emph{POLARITY} ~\cite{ReviewPangLee2004}). More extensively, two object recognition datasets have been taken into account (PASCAL VOC 2007-2012~\cite{pascal-voc-2007,pascal-voc-2012}). Results show that the proposed approach represents the most robust algorithm, which achieves the highest level of performance across many different domains and challenging scenarios.

The rest of the paper is organized as follows: Sec.~\ref{sec:soa} illustrates the related literature, mostly focusing on the comparative approaches we consider in this study. Sec.~\ref{sec:method} details the proposed approach, also giving a formal justification and interpretation based on absorbing Markov chain (Sec.~\ref{sec:markovProcesses}). Extensive experiments are reported in Sec.~\ref{sec:exp}, and, finally, in Sec.~\ref{sec:conc}, conclusions are given, and future perspectives are envisaged.

\vspace{-0.2cm}
%
%
\section{Related Work}\label{sec:soa}\vspace{-0.2cm}

Since the mid-1990s, few domains used more than 20 features. The situation has changed considerably in the past few years and most papers explore domains with hundreds to tens of thousands of features. New approaches were proposed to address these challenging tasks involving many irrelevant and redundant variables and often comparably few training examples. 
Typically, FS techniques are partitioned into three classes ~\cite{Guyon:2002}: \textit{Filters}, \textit{Wrappers} and \textit{Embedded} methods. The proposed approach is a filter method, which analyzes intrinsic properties of data, ignoring the type of classifier. Conversely, wrappers use classifiers to score a given subset of features, and embedded methods inject the selection process directly into the learning process of the classification framework.

Among the most used filter-based strategies, \emph{Relief-F}~\cite{liu2008} is an iterative, randomized, and supervised approach that estimates the quality of the features according to how well their values differentiate data samples that are near to each other. Another effective yet fast filter method is the \textit{Fisher} method~\cite{Quanquanjournals}, which computes a score for a feature as the ratio of inter-class separation and intra-class variance, where features are evaluated independently. A Mutual Information based approach (\emph{MI}) is proposed in \cite{Hutter:02feature}. MI considers as a selection criterion the mutual information between the distribution of the values of a given feature and the membership to a particular class. Even in the last case, features are evaluated independently, and the final feature selection occurs by aggregating the $m$ top ranked ones. In unsupervised learning scenarios, a widely used method is the Laplacian Score (LS)~\cite{HCN05a}, where the importance of a feature is evaluated by its power of locality preserving. In order to model the local geometric structure, this method constructs a nearest neighbor graph. LS algorithm seeks those features that respect this graph structure. The unsupervised feature selection for multi-cluster data is denoted MCFS in \cite{Cai:2010}, which selects those features such that the multi-cluster structure of the data can be best preserved. \cite{Yang:2011} proposed a L2,1-norm regularized discriminative feature selection for unsupervised learning (UDFS) which selects the most discriminative feature subset from the whole feature set in batch mode. Feature selection and kernel learning for local learning-based clustering (LLCFS)~\cite{zeng2011feature} associates a weight to each feature and incorporates it into the built-in regularization of the LLC algorithm to take into account the relevance of each feature for the clustering. In the experiments, we also compare our approach against the unsupervised graph-based filter method dubbed Inf-FS~\cite{Roffo:InfFS:2015}. In the Inf-FS formulation, each feature is a node in the graph, a path is a selection of features, and the higher the centrality score, the most important (or most different) the feature. Another widely used FS method is SVM-RFE (RFE)~\cite{Guyon:2002}, which is a wrapper method that selects features in a sequential, backward elimination manner, ranking high a feature if it strongly separates the samples by means of a linear SVM.  Finally, for the embedded methods, the \emph{feature selection via concave minimization} (\emph{FSV})~\cite{Bradley98featureselection} is a popular FS strategy, where the selection process is injected \emph{into} the training of an SVM by a linear programming technique. For further information, please see Tab. \ref{table:compmethods}. 
\vspace{-0.2cm}

%
%
\section{Our Approach}\label{sec:method}\vspace{-0.2cm}

Given a training set $X$ represented as a set of feature distributions $X = \{ \vec x_1, ..., \vec x_n  \}$, where each $m \times 1$ vector $\vec x_i$ is the distribution of the values assumed by the $i^{th}$ feature with regards to the $m$ samples, we build an undirected graph $G$, where nodes correspond to features and edges model relationships among pairs of nodes. Let the adjacency matrix $A$ associated to $G$ defining the nature of the weighted edges: each element $a_{ij}$ of $A$, $1\leq i,j \leq n$, models pairwise relationships between the features. Each weight represents the likelihood that features $\vec x_i$ and $\vec x_j$ are good candidates. Weights can be associated to a binary function of the graph nodes:
\begin{equation}\label{eq:partzero}
  a_{ij} = 	\varphi(\vec x_i,\vec x_j),
\end{equation}
where $\varphi(\cdot,\cdot)$ is a real-valued potential function learned by the proposed approach in a PLSA-inspired framework. The learning framework models the probability of each co-occurrence in $\vec x_i, \vec x_j$ as a mixture of conditionally independent multinomial distributions, where parameters are learnt using the EM algorithm.  
Given the weighted graph $G$, the proposed approach analyses subsets of features as paths connecting them. The cost of each path is given by the joint probability of all the nodes belonging to it. The method exploits the convergence property of the power series of matrices as in \cite{Roffo:InfFS:2015}, and evaluates in an elegant fashion the relevance of each feature with respect to all the other ones taken together. For this reason, we dub our approach \emph{infinite latent feature selection} (ILFS).

%
%
\subsection{Discriminative Quantization process}\label{sec:DQ}\vspace{-0.2cm}

Since the amount of possible distinct values in $\vec x_i$ is huge, we map this large set of values to a countable smaller set, hereinafter referred to as set of \emph{tokens}. Tokens are the words of our dictionary of features. Thus, each feature will be represented by a new low-dimensional vocabulary of meaningful tokens. The way used to assign each value to a specific token is based on a quantization process, we called \emph{discriminative quantization} (DQ). The rationale behind the DQ process is to take into account how well a given feature is representative of a class before performing the many-to-few mapping.

Firstly, the Fisher criterion is used to compute a scoring vector $\Phi = [\cdot, ..., \cdot]$ which takes into account both means and standard deviations of the classes, for each sample and feature. In binary classification scenarios, this is given by
\begin{equation}\label{eq:priors1}
      \Phi = \frac{1}{\mathcal{Z}} \Big[\frac{(s - \mu_1)^2}{\sigma_1^2 + \sigma_2^2} ,  \frac{(s - \mu_2)^2}{\sigma_1^2 + \sigma_2^2}  \Big],    
\end{equation}
    
where $s$ is a sample from the $i^{th}$ feature $\vec x_i$, $\mu_k$ and $\sigma_k$ denote the mean and standard deviation of class $k$, respectively. A normalization factor $\mathcal{Z}$ is introduced to ensure that the scores are a valid distribution over both classes. A natural generalization of these scores into a multi-class framework is given by
\begin{equation}\label{eq:priors2}
	\Phi = \frac{1}{\mathcal{Z}} \Big[\frac{ (s - \mu_{1})^2 }{\sum_{k=1}^K \sigma_{k}^2}, ..., \frac{ (s - \mu_{K})^2 }{\sum_{k=1}^K \sigma_{k}^2}\Big], \forall_{k \in K}
\end{equation}
where $K$ is the number of classes, $s$ is a single sample from the $i^{th}$ feature. Therefore, considering all the samples, $\Phi$ results to be a $m \times K$ matrix. 

Now, let us assume that the sample $s$ belongs to class $k$. If $\vec x_i$ is a strong discriminant feature, $s$ will \emph{score high at $\Phi_k$}. Then, we derive our priors $\pi$ by extracting $\Phi$ scores for each feature according to the ground truth as follows:
\[
    \pi = diag(\Phi Y)
\]
where $Y$ is the 1-of-K representation of the ground truth. It is a particularly convenient representation where the class labels are represented by K-dimensional vectors in which one of the elements equals 1, and all remaining elements equal 0. As a result, $\pi \in [0,1]$ is a $1 \times m$ vector containing a score for each element of a particular feature $i$. It takes into account how well each element is represented by the feature $i$ according to Eq.\ref{eq:priors2}. 

Finally, quantization is performed. The first step is to divide the entire range of values $[0,1]$ into a series of $\mathcal{T}$ intervals (i.e., we use $\mathcal{T}=6$ in this work: interval $1$ corresponds to not-well-represented samples, and interval $6$ is associated to well-represented samples). Secondly, we assign a token to values falling into each interval. Given the outcomes of the DQ process, we obtain a meaningful new representation of our training data $X$ in the form of $F = \{ f_1, ..., f_n  \}$, where each feature is described by a vocabulary of few tokens. In other words, the derived feature representation $f_i$ comes from $x_i$ where each value is assigned to a token $\mathcal{T}$. According to this formulation, a strong discriminative feature will be intuitively associated to a descriptor $f_i$ containing many relatively large tokens (e.g., $5,6$) rather than small ones (e.g., $1,2$).
\begin{figure}[!]
\centering
\includegraphics[width=0.4\textwidth]{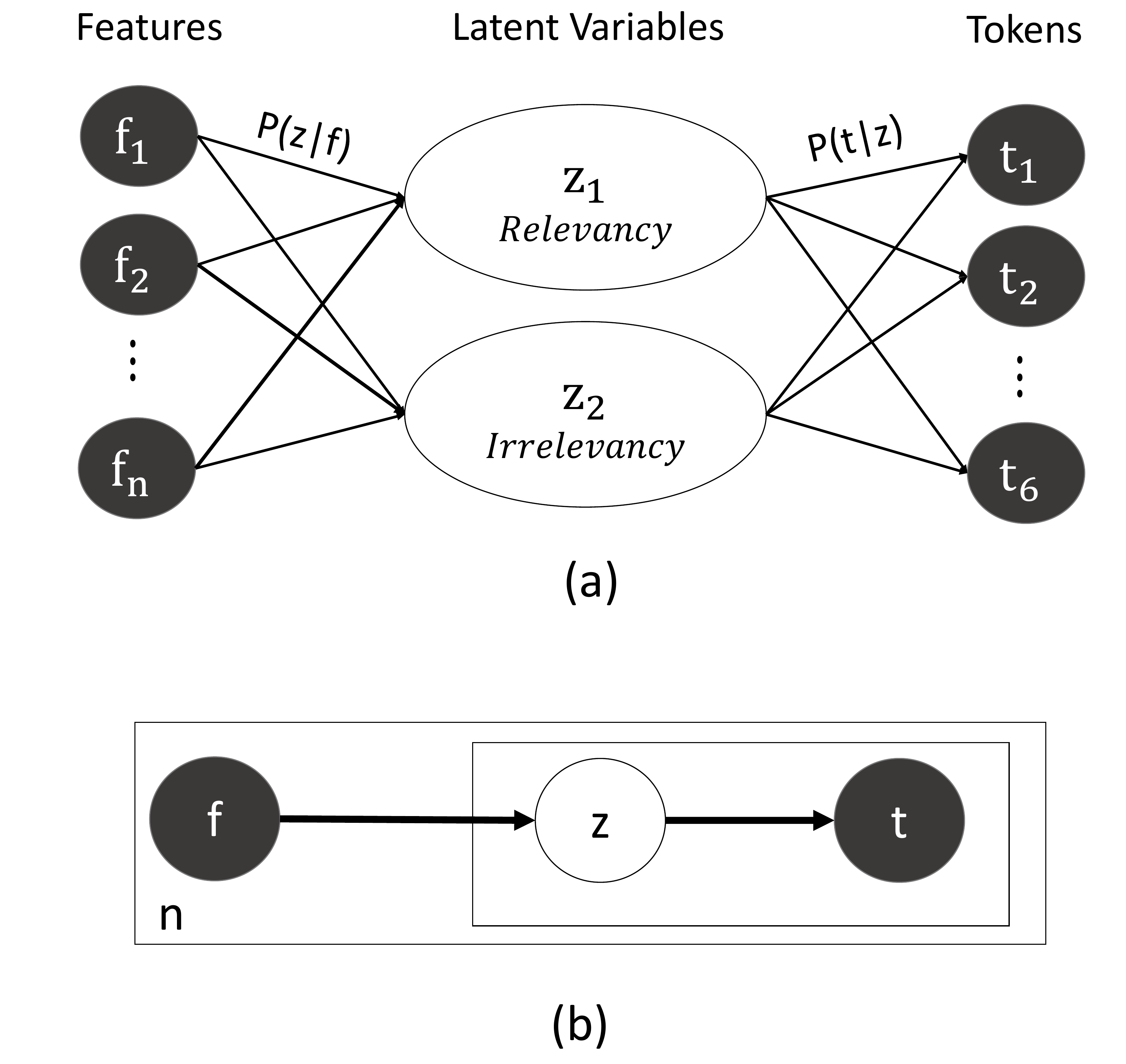}
\caption{Illustration of the general structure of the model. (a) The intermediate layer of latent topics that links the features and the tokens. (b) The graphical model using plate representation. }\label{fig:model}
\end{figure}

%
%
\subsection{From co-occurrences to graph weighting}\label{sec:LEARN}\vspace{-0.2cm}

Weighting the graph according to the nodes discriminatory power has a great influence on the quality of the ranking process. We designed a framework to automatically perform the graph weighting from training data, such that the learnt parameters can be used to sort features according to their degrees of relevance or importance.

Our solution is based on a variation of the PLSA ~\cite{hofmann1999probabilistic} technique, that considers co-occurrences of tokens and features, $\langle t,f \rangle$, to model the probability of each co-occurrence as a mixture of conditionally independent multinomial distributions.

In order to better understand the intuition behind the proposed model, we need to make some assumptions. We assume that a feature consists of only two topics representing the two main latent variables of any feature selection algorithms: \textit{Relevancy} and \textit{Irrelevancy}. Therefore, we introduce an unobserved class variable $\mathcal{Z} = \{z_1, z_2\}$ obtaining a latent variable model for co-occurrence tokens. As a result,
there is a distribution $P(z|f)$ over the fixed number of topics for each feature $f$. Similarly, original PLSA
model does not have the explicit specification of this distribution but it is indeed a multinomial distribution
where $P(z|f)$ represents the probability that topic $z$ appears in feature $f$. Fig.~\ref{fig:model}.(\textit{a}) shows the general structure of the model, each feature can be represented as a mixture of concepts (Relevant/Irrelevant) weighted by the probability $P(z|f)$ and each token expresses a topic with probability $P(t|z)$. Fig.~\ref{fig:model}.(\textit{b}) describes the generative process for each of the $n$ features in the set by using plate representation.
We can write the probability a token $t$ appearing in feature $f$ as follows:
\[
    P(t|f) = P(t | z_1)P(z_1 | f) + P(t | z_2)P(z_2 | f).
\]
By replacing this for any feature in the set $F$ we obtain,
\[
    P(f) = \prod_t \Big\{  P(t | z_1)P(z_1 | f) + P(t | z_2)P(z_2 | f)  \Big\}.
\]
The unknown parameters of this model are $P(t|z)$ and $P(z|f)$. As for PLSA, we derived the equation for computing these parameters by maximum likelihood. The log-likelihood function is given by
\[
    \mathcal{L} =  \sum_f \sum_t Q(f,t) \log[P(t|f)]
\]
where $Q(f,t)$ is the number of times token $t$ appearing in feature $f$. The EM algorithm is used to compute optimal parameters. The E-step is given by
\[
    P(z|f,t) = \frac{P(z)P(f|z)P(t|z)}{P(z_1)P(f|z_1)P(t|z_1)+P(z_2)P(f|z_2)P(t|z_2)},
\]
and the M-step is given by
\[
    P(t|z) = \frac{\sum_f Q(f,t)P(z|f,t)}{\sum_{f,t'} Q(f,t')P(z|f,t')},
\]
\[
    P(f|z) = \frac{\sum_t Q(f,t)P(z|f,t)}{\sum_{f',t} Q(f',t)P(z|f',t)},
\]
\[
    P(z) = \frac{\sum_{f,t} Q(f,t)P(z|f,t)}{\sum_{f,t} Q(f,t)}.
\]

The responsibility for assigning the ``condition of being relevant" to features lies to a great extent with the unobserved class variable $Z$. In particular, we initialize the model priors $P(t|z)$ in order to link $z_1$ to the abstract topic of \textit{Relevancy}, and hence $z_2$ to \textit{Irrelevancy}. By construction we limited the range of the tokens to values between $1$ and $6$ (see Sec.\ref{sec:DQ}), with $1$ that behaves the same way as being the lowest rating for a sample of a particular feature, and $6$ being the highest quality. As a result, a natural way to initialize these priors is to generate a pair of linearly spaced vectors assigning a higher probability $P( t' |Z = z_1)$ for those tokens $t'$ which score higher, and consequently the opposite for $P( t' |Z = z_2)$. 

Finally, the graph can be weighted by the estimated probability distribution $P(Z = z_1 | f)$. According to Eq.\ref{eq:partzero}, each element $a_ij$ of the adjacency matrix is the joint probability that the abstract topic of relevancy appears in feature $f_i$ and $f_j$, namely:
\begin{equation}\label{eq:weigthing}
  a_{ij} = 	\varphi(\vec x_i,\vec x_j) = P(Z = z_1 | f_i) P(Z = z_1 | f_j),
\end{equation}
where mixing weights $P(Z = z_1 | f_i)$ and $P(Z = z_1 | f_j)$ are conditionally independent. Indeed, knowledge of whether $P(Z = z_1 | f_i)$ occurs provides no information on the likelihood of $P(Z = z_1 | f_j)$ occurring, and knowledge of whether $P(Z = z_1 | f_j)$ occurs provides no information on the likelihood of $P(Z = z_1 | f_i)$ occurring.

%
%
\subsection{Probabilistic Infinite Feature Selection}\label{sec:infFS}\vspace{-0.2cm}

Let $\gamma = \{ v_{0}=i,v_{1}, ..., v_{l-1}, v_{l}=j \}$ denote a path of length $l$ between nodes $i$ and $j$, that is, features $\vec x_i$ and $\vec x_j$, through other nodes $v_{1},...,v_{l-1}$. For simplicity, suppose that the length $l$ of the path is lower than the total number of nodes $n$ in the graph. In this setting, a path is simply a subset of the available features/nodes that come into play. Moreover, the network is characterized by walk structure~\cite{Borgatti2005}, where nodes and edges can be visited multiple times. 

We can then estimate the joint probability that $\gamma$ is a good subset of features as
\begin{equation}\label{eq:two}
\mathcal{P}_{\gamma } = \prod_{k=0}^{l-1} a_{v_{k},v_{k+1}}.
\end{equation}

Let us define the set $\mathbb{P}_{i,j}^l$ as containing all the paths of length $l$ between $i$ and $j$; to account for the energy of all the paths of length $l$, we sum them as follows:
\begin{equation}\label{eq:three}
C_{l}(i,j) =\sum_{\gamma \in \mathbb{P}_{i,j}^l }  \mathcal{P}_{\gamma},
\end{equation}
which, following standard matrix algebra, gives:
$$ C_{l}(i,j) = A^{l}(i,j) , $$
that is, the adjacency matrix $A$ elevated by $l$.

However, we want to consider all the possible paths of any length in the graph, which turns out to be the same as considering all the the possible subsets of features of any cardinality. Therefore, extending the path length to infinity implies that we have to calculate the geometric series of matrix $A$
\begin{equation}\label{eq:fourPOINTfive2}
 \hat C = \sum _{l=1}^{\infty} A^{l}.
\end{equation}
Summing infinite $A^l$ terms brings divergence. Therefore, regularization is needed. Regularization is used to assign a consistent value for the sum of a possibly divergent series. Among the different forms of regularization~\cite{Bergshoeff,Graham:1994}, we use a simple generating function for the $l$-path as
\begin{equation}\label{eq:fourPOINTfive2}
 \check{C} = \sum _{l=1}^{\infty} r^l A^{l},
\end{equation}
where $\textit{r}$ is a real-valued regularization factor, and $\textit{r}^{l}$ can be interpreted as the weight for paths of length $l$. Thus, for appropriate choices of $\textit{r}$, it is ensured that the infinite sum converges. 
From an algebraic point of view, $\check{C}$ can be efficiently computed by using the convergence property of the geometric power series of a matrix~\cite{HubHub01}:
\begin{equation}\label{eq:six}
          \check{C} = (I - \textit{r}A)^{-1} - I,
\end{equation}
Matrix $\check{C}$ encodes all the information about the goodness of our set of features.
We can obtain final scores for each node simply by marginalizing this quantity:
\begin{equation}\label{eq:seven}
          \check{c}(i) = [\check{C} \textbf{e}]_{i},
\end{equation}
where $\textbf{e}$ indicates a 1D array of ones. Ranking in decreasing order the $\check{c}(i)$ scores gives the output of the algorithm: a ranked list of features where the most discriminative and relevant features are positioned at the top of the list. The gist of the ILFS is to provide a score of importance for each feature as a function of the importance of its neighbors.

 \begin{table*}[!]
\small
\centering
\resizebox{0.90\textwidth}{!}{%
\begin{tabular}{l c c c c c c c c c }
\hline \hline
\textbf{Dataset} & \textbf{Ref.} & \textbf{\#Samples} & \textbf{\#Classes} & \textbf{\#Feat.}  & \textbf{\emph{few train}}& \textbf{\emph{unbal. (+/-)}} & \textbf{\emph{overlap}} & \textbf{\emph{noise}} & \textbf{\emph{sparse}}\\\hline
GINA &~\cite{GINA} & 3153 &2& 970   &   & (1,5K/1,6K) & X & & \\ 
DEXTER &~\cite{NIPS2003} &  2600 &2& 20K   &   & (1,3K/1,3K) & X &  & X\\
\emph{POLARITY} &~\cite{ReviewPangLee2004} & 2K &2& 3K   &   & (1K/1K)  &  & & X\\
\hline
\emph{COLON} &~\cite{alon} & 62 &2& 2K   &  X & (40/22) & \emph{n.s.}& X  &  \\
\emph{LEUKEMIA} &~\cite{Golub99} & 72 &2& 7129   &  X & (47/25) &\emph{n.s.} & X & \\
\emph{PROSTATE} &~\cite{citeulike:1624492}  & 102 & 2 & 6033   &  X & (50/52)&\emph{n.s.}& &  \\
\emph{LYMPHOMA} &~\cite{Golub99} & 45 &2& 4026   &  X & (23/22) & \emph{n.s.}& &  \\
\emph{LUNG} &~\cite{Gordon02} & 181 &2& 12533   &  X & (31/150) & \emph{n.s.}& X & \\
\hline \vspace{0.02cm}
VOC 2007 &~\cite{pascal-voc-2007} & ~10K &20& 4096   & & X & X & X & \\
VOC 2012 &~\cite{pascal-voc-2012} & ~20K &20& 4096   & & X & X & X & \\
\hline
\end{tabular}}
\caption{Datasets and the challenges for the feature selection scenario. The abbreviation \emph{n.s.} stands for \emph{not specified} (for example, in the DNA microarray datasets, any information on class overlap is given in advance).  }
\label{table:ch5datasets}
\end{table*}

\subsection{Markov chains and random walks}\label{sec:markovProcesses}\vspace{-0.1cm}

This section provides a probabilistic interpretation of the proposed algorithm based on Absorbing Random Walks. 
Here, we reformulate the problem in terms of Markov chains and random walks. The set of nodes in a Markov chain are called \textit{states} and each move is called a \textit{step}. 
Let $T$ be the \textit{matrix of transition probabilities}, or the \textit{transition matrix} of the Markov chain. If the chain is currently in state $v_i$, then it moves to state $v_j$ at the next step with a probability denoted by $t_{ij}$, and this probability does not depend upon which states the chain was in before the current state. The probabilities $t_{ij}$ are called transition probabilities.
The process can remain in the state it is in, and this occurs with probability $t_{ii}$. 
An absorbing Markov chain is a special Markov chain which has absorbing states, i.e., states which once reached cannot be transitioned out of (i.e., $t_{ii} = 1$). A Markov chain is absorbing if it has at least one absorbing state, and if from every state it is possible to go to an absorbing state in a finite number of steps. In an absorbing Markov chain, a state that is not absorbing is called transient.
The transition matrix for any absorbing chain can be written in the \textit{canonical} form
\[
T =
  \begin{bmatrix}
    \textbf{I} & \textbf{0}  \\
    R & A
  \end{bmatrix}
\]
where $R$ is the rectangular submatrix giving transition probabilities from non-absorbing to absorbing states, $A$ is the square submatrix giving these probabilities from non-absorbing to non-absorbing states, $\textbf{I}$ is an identity matrix, and \textbf{0} is a rectangular matrix of zeros.

Note that $R$ and $\textbf{0}$ are not necessarily square. More precisely, if there are $m$ absorbing states
and $n$ non-absorbing states, then $R$ is $n \times m$, $A$ is $n \times n$ , \textbf{I} is $m \times m$, and \textbf{0} is $m \times n$. Iterated multiplication of the $T$ matrix yields

\[
T^2 =
  \begin{bmatrix}
    \textbf{I} & \textbf{0}  \\
    R & A
  \end{bmatrix}
\begin{bmatrix}
    \textbf{I} & \textbf{0}  \\
    R & A
  \end{bmatrix}
=
\begin{bmatrix}
    \textbf{I} & \textbf{0}  \\
    R+AR & A^2
  \end{bmatrix}
\]
\[
T^3 =
 \begin{bmatrix}
    \textbf{I} & \textbf{0}  \\
    R+AR & A^2
  \end{bmatrix}
\begin{bmatrix}
    \textbf{I} & \textbf{0}  \\
    R & A
  \end{bmatrix}
=
\begin{bmatrix}
    \textbf{I} & \textbf{0}  \\
    R+AR+A^2R & A^3
  \end{bmatrix}
\]
and hence by induction we obtain
\[
T^l =
\begin{bmatrix}
    \textbf{I} & \textbf{0}  \\
   (\textbf{I}+A+A^2+...+A^{l-1})R & A^l
  \end{bmatrix}
\]
The preceding example illustrates the general result that $A^l \to 0$ as $l \to \infty$. Thus
\[
T^{\infty} =
\begin{bmatrix}
    \textbf{I} & \textbf{0}  \\
   CR & \textbf{0}
  \end{bmatrix}
\]
where the matrix
\[
C = \textbf{I} + A + A^2 + ... + A^{\infty}= (I-A)^{-1}
\]
is called the \textit{fundamental matrix} for the absorbing chain. Note that $C$, which is a square matrix with rows and columns corresponding to the non-absorbing states, is derived in the same way of Eq.\ref{eq:six}. $C(i, j)$ is the expected number of periods that the chain spends in the $j^{th}$ non-absorbing state given that the chain began in the $i^{th}$ non-absorbing state. Perhaps this interpretation comes from the specification of the matrix $C$ as the infinite sum, since $A^l(i, j)$ is the probability that the process which began in the $i^{th}$ non-absorbing state will occupy the $j^{th}$ non-absorbing state in period $l$. However, $A^l(i, j)$ can also be understood as the expected proportion of period $l$ spent in the $j^{th}$ state. Summing over all time periods $l$, we thus obtain the total number of periods that the chain is expected to occupy the $j^{th}$ state.

\vspace{-0.25cm}
%
%
\section{Experiments and Results}\label{sec:exp}\vspace{-0.2cm}

This section has three main goals. The first goal is to evaluate the robustness of the proposed method, by choosing datasets spanning over a variety of domains and difficulties. For example, we consider the problems of dealing with few training samples and many features (few train in Tab.~\ref{table:ch5datasets}), sparse or dense dataset, unbalanced classes (unbalanced), or classes that severely overlap (overlap), or whose samples are noisy (noise) due to: a) complex scenes where the object to be classified is located (as in the PASCAL VOC series) or b) many outliers (as in the genetic datasets, where samples are often contaminated, that is, artifacts are present into the data during the acquisition of the samples). The second goal is to analyze and empirically clarify how well important features are ranked high by the ILFS. We also include several comparative algorithms from recent literature, including filters, wrappers, and embedded methods. The last goal is to assess the reliability and validity of our research results. We present results obtained from more than $550$ different tests, evaluating if the difference in performance is statistically significant by means of a set of Student's t-test and binomial cumulative distribution functions.

\noindent
\textbf{Comparative approaches and complexity} \\
Tab. \ref{table:compmethods} lists the methods compared, where we note their \emph{type} (\emph{f} = filters, \emph{w} = wrappers, \emph{e} = embedded methods), and their \emph{class} ( \emph{s} = supervised or \emph{u} = unsupervised, i.e., using or not using the labels associated with the training samples in the ranking operation). Additionally, we report their computational complexity (if it is documented in the literature). The complexity of our approach is $\mathcal{O}(n^{2.37}+in+T+C)$, the matrix inversion for a $n\times n$ matrix requires $O(n^{2.37})$ \cite{matrixInversion2}, and the second term $O(in+T+C)$ comes from the estimate of $P(z|f)$ through PLSA; hidden constants are the number of latent variables ($Z=2$) and the number of tokens used ($\mathcal{T}=6$). 
Finally, Tab. \ref{table:compmethods} reports the execution time of each method when applied to a randomly generated dataset consisting of $2$ classes, $10k$ samples, and $5k$ features (features follow a uniform distribution - range [0,1000]), on an Intel i7 CPU 3.4GHz, 16.0 GB of RAM, using MATLAB 2016b. 

\begin{table}[!]
\small
\centering
\resizebox{0.45\textwidth}{!}{%
\begin{tabular}{| c |p{1.65cm} |C{0.5cm}| C{0.5cm} |C{2.9cm}| C{1.4cm}|}
\hline
\textbf{ID} &\textbf{Acronym} &   \textbf{\small{Type}} & \textbf{\small{Cl.}} & \textbf{Comp. Complexity}  & \textbf{\small{Exec.Time}}\\\hline
1 & CFS \cite{Guyon:2002} &f&u& $\mathcal{O}(\frac{n^2}{2}T)$  & 2 \\\hline
2 & Fisher~\cite{Quanquanjournals}   &f&s& $\mathcal{O}(Tn)$  & 1 \\\hline
3 & FSV~\cite{Bradley98featureselection} & e& s& \small{$\mathcal{O}(T^2n^2)$}&2985 \\\hline
4 & LLCFS \cite{zeng2011feature}  &f&u& N/A  & 2934\\\hline
5 & LS \cite{HCN05a} &f&u& N/A  & 455\\\hline
6 & MCFS \cite{Cai:2010}&f&u&  N/A  & 10 \\\hline
7 & MI~\cite{Hutter:02feature} &f& s&$\sim\mathcal{O}( n^2 T^2)$& 7\\\hline
8 & Relief-F~\cite{liu2008} &f&s& $\mathcal{O}(iTnC)$  & 2024\\\hline
9 & RFE \cite{Guyon:2002} & w& s& \small{$\mathcal{O}(T^2 n log_2n )$} &91799\\\hline
10& UDFS \cite{Yang:2011}  &f&u& N/A & 1954\\\hline
11& Inf-FS \cite{Roffo:InfFS:2015}  &f&u& $\mathcal{O}(n^{2.37}(1+T))$ & 12\\\hline
12& Ours &f&s& $\mathcal{O}(n^{2.37}+in+T+C)$ & 7 \\\hline
\end{tabular}}
\caption{Feature selection approaches considered in the experiments \cite{Roffo2017b,roffo2016feature}. The table reports their \emph{Type}, class (\emph{Cl.}), complexity (\emph{Compl.}), and execution times in seconds (\emph{Exec.Time}). As for the complexity, $T$ is the number of samples, $n$ is the number of initial features, $i$ is the number of iterations in the case of iterative algorithms, and $C$ is the number of classes.}
\label{table:compmethods}
\end{table}

\begin{table*}[t!]
\small
\centering
\resizebox{1\textwidth}{!}{%
\begin{tabular}{|C{0.44cm}|C{0.5cm} |C{0.5cm}|C{0.5cm} |C{0.5cm}| C{0.5cm}|C{0.5cm} |  C{0.5cm} |C{0.5cm}| C{0.5cm}| C{0.5cm}| C{0.5cm}| C{0.52cm}| C{0.01cm}| C{0.5cm}| C{0.5cm}| C{0.5cm}| C{0.52cm} |C{0.5cm} |C{0.5cm}| C{0.5cm}|C{0.5cm}| C{0.5cm}| C{0.5cm}| C{0.5cm} | C{0.5cm} |  }
\hline
\multicolumn{26}{|c|}{\textbf{The PASCAL Visual Object Classes (VOC)}} \\
\hline
 & \multicolumn{12}{c|}{ \textbf{VOC 2007}} & &\multicolumn{12}{c|}{ \textbf{VOC 2012}} \\
\hline
& \tiny{CFS}  & \tiny{Fisher} &\tiny{FSV} &\tiny{LLCFS} & \tiny{LS} & \tiny{MCFS} & \scriptsize{MI} & \tiny{ReliefF} &\tiny{RFE} & \tiny{UDFS} & \tiny{Inf-FS} &\scriptsize{\textbf{Ours}}  & & \tiny{CFS} &\tiny{Fisher} &\tiny{FSV} & \tiny{LLCFS} & \tiny{LS} & \tiny{MCFS}& \tiny{MI} & \tiny{ReliefF} &\tiny{RFE} & \tiny{UDFS} & \tiny{Inf-FS} &\scriptsize{\textbf{Ours}}  \\
 \hline
\centering
	$ \vcenter{\includegraphics[scale=0.1]{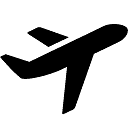}} $ & \scriptsize{90.72} & \scriptsize{\textbf{92.67}} & \scriptsize{91.57} & \scriptsize{91.32} & \scriptsize{91.43} & \scriptsize{91.00} & \scriptsize{92.46} & \scriptsize{90.30} & \scriptsize{91.44} & \scriptsize{91.98} &\scriptsize{91.37}& \scriptsize{91.75} &
 & 
\scriptsize{96.83} & \scriptsize{96.97} & \scriptsize{97.20} & \scriptsize{97.70} & \scriptsize{97.32} & \scriptsize{97.30} & \scriptsize{\textbf{97.35}} & \scriptsize{96.54}& \scriptsize{96.95} & \scriptsize{96.84} &\scriptsize{ 96.11}& \scriptsize{97.05} 
\\
\hline 
\centering
	$ \vcenter{\includegraphics[scale=0.1]{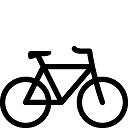}} $ & \scriptsize{87.09} & \scriptsize{86.76} & \scriptsize{84.91} & \scriptsize{86.42} & \scriptsize{87.25} & \scriptsize{87.44} & \scriptsize{\textbf{87.79}} & \scriptsize{85.66} & \scriptsize{85.00} & \scriptsize{87.57} &\scriptsize{87.21 }&\scriptsize{87.60} &
 & 
\scriptsize{82.01} & \scriptsize{82.72} & \scriptsize{82.19} & \scriptsize{82.52} & \scriptsize{82.44} & \scriptsize{82.64} & \scriptsize{82.69} & \scriptsize{81.42} & \scriptsize{78.68} & \scriptsize{82.52} &\scriptsize{ 79.05}& \scriptsize{\textbf{82.83}}
\\
\hline 
\centering
	$ \vcenter{\includegraphics[scale=0.1]{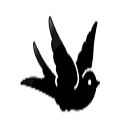}} $ & \scriptsize{89.72} & \scriptsize{90.17} & \scriptsize{89.51} & \scriptsize{89.11} & \scriptsize{89.74} & \scriptsize{90.23} & \scriptsize{88.75} & \scriptsize{89.20} & \scriptsize{88.61} & \scriptsize{89.12} &\scriptsize{ 89.25}& \scriptsize{\textbf{90.25}} &
 & 
\scriptsize{89.75} & \scriptsize{90.21} & \scriptsize{89.84} & \scriptsize{89.91} & \scriptsize{89.80} & \scriptsize{90.07} & \scriptsize{\textbf{90.28}} & \scriptsize{89.19} & \scriptsize{88.56} & \scriptsize{89.81} &\scriptsize{88.44 }& \scriptsize{89.44}
\\
\hline 
\centering
	$ \vcenter{\includegraphics[scale=0.1]{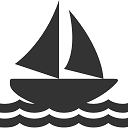}} $ & \scriptsize{88.28} & \scriptsize{88.33} & \scriptsize{\textbf{88.83}} & \scriptsize{88.32} & \scriptsize{88.45} & \scriptsize{87.60} & \scriptsize{88.11} & \scriptsize{88.18} & \scriptsize{87.51} & \scriptsize{88.28} &\scriptsize{ 88.41}& \scriptsize{88.57} &
 & 
\scriptsize{89.32} & \scriptsize{90.00} & \scriptsize{89.88} & \scriptsize{89.37} & \scriptsize{89.80} & \scriptsize{89.60} & \scriptsize{89.96} & \scriptsize{89.09} & \scriptsize{87.39} & \scriptsize{89.39} &\scriptsize{ 88.05}&\scriptsize{\textbf{90.20}}
\\
\hline 
\centering
	$ \vcenter{\includegraphics[scale=0.1]{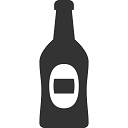}} $ & \scriptsize{56.45} & \scriptsize{56.06} & \scriptsize{56.27} & \scriptsize{54.44} & \scriptsize{55.53} & \scriptsize{54.83} & \scriptsize{55.80} & \scriptsize{54.51} & \scriptsize{50.35} & \scriptsize{57.84} &\scriptsize{54.63 }& \scriptsize{56.18} &
 & 
\scriptsize{60.02} & \scriptsize{60.99} & \scriptsize{60.61} & \scriptsize{60.45} & \scriptsize{60.18} & \scriptsize{60.81} & \scriptsize{\textbf{62.21}} & \scriptsize{57.93} & \scriptsize{50.91} & \scriptsize{61.31} &\scriptsize{56.18 }& \scriptsize{61.47}
\\
\hline 
\centering
	$ \vcenter{\includegraphics[scale=0.1]{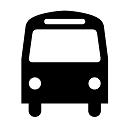}} $ & \scriptsize{81.71} & \scriptsize{81.74} & \scriptsize{82.07} & \scriptsize{81.50} & \scriptsize{81.21} & \scriptsize{81.76} & \scriptsize{82.16} & \scriptsize{80.97} & \scriptsize{80.12} & \scriptsize{81.28} &\scriptsize{ 81.20}& \scriptsize{\textbf{83.02}} &
 & 
\scriptsize{88.05} & \scriptsize{88.66} & \scriptsize{88.46} & \scriptsize{\textbf{89.55}} & \scriptsize{88.36} & \scriptsize{88.47} & \scriptsize{88.69} & \scriptsize{87.42} & \scriptsize{88.16} & \scriptsize{88.69} &\scriptsize{ 86.51}& \scriptsize{89.36}
\\
\hline 
\centering
	$ \vcenter{\includegraphics[scale=0.1]{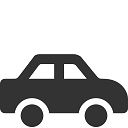}} $ & \scriptsize{86.97} & \scriptsize{87.32} & \scriptsize{\textbf{87.77}} & \scriptsize{87.28} & \scriptsize{87.09} & \scriptsize{87.13} & \scriptsize{87.47} & \scriptsize{87.93} & \scriptsize{85.52} & \scriptsize{87.71} &\scriptsize{87.47 }& \scriptsize{87.23} &
 & 
\scriptsize{81.42} & \scriptsize{\textbf{81.91}} & \scriptsize{81.62} & \scriptsize{81.31} & \scriptsize{81.26} & \scriptsize{81.67} & \scriptsize{81.77} & \scriptsize{80.30} & \scriptsize{73.98} & \scriptsize{81.02} &\scriptsize{78.80 }& \scriptsize{81.74}
\\
\hline 
\centering
	$ \vcenter{\includegraphics[scale=0.1]{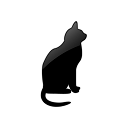}} $ & \scriptsize{86.61} & \scriptsize{87.21} & \scriptsize{87.44} & \scriptsize{87.49} & \scriptsize{\textbf{88.06}} & \scriptsize{87.28} & \scriptsize{86.85} & \scriptsize{86.82} & \scriptsize{86.57} & \scriptsize{87.46} &\scriptsize{ 87.61}& \scriptsize{86.61} &
 & 
\scriptsize{93.10} & \scriptsize{93.04} & \scriptsize{93.24} & \scriptsize{92.83} & \scriptsize{93.28} & \scriptsize{\textbf{93.43}} & \scriptsize{93.14} & \scriptsize{92.96} & \scriptsize{92.07} & \scriptsize{93.16} &\scriptsize{91.24 }&\scriptsize{92.83}
\\
\hline 
\centering
	$ \vcenter{\includegraphics[scale=0.1]{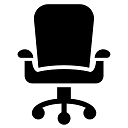}} $ & \scriptsize{67.05} & \scriptsize{67.19} & \scriptsize{63.50} & \scriptsize{67.25} & \scriptsize{67.53} & \scriptsize{67.14} & \scriptsize{67.35} & \scriptsize{64.74} & \scriptsize{59.34} & \scriptsize{66.93} &\scriptsize{\textbf{67.61} }&\scriptsize{66.96} &
 & 
\scriptsize{71.04} & \scriptsize{\textbf{72.44}} & \scriptsize{70.46} & \scriptsize{71.40} & \scriptsize{72.29} & \scriptsize{71.70} & \scriptsize{71.60} & \scriptsize{69.72} & \scriptsize{59.31} & \scriptsize{72.03} &\scriptsize{67.42 }&\scriptsize{71.89}
\\
\hline 
\centering
	$ \vcenter{\includegraphics[scale=0.1]{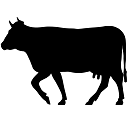}} $ & \scriptsize{75.79} & \scriptsize{76.38} & \scriptsize{74.94} & \scriptsize{75.47} & \scriptsize{76.36} & \scriptsize{76.16} & \scriptsize{76.31} & \scriptsize{73.84} & \scriptsize{73.84} & \scriptsize{75.16} &\scriptsize{\textbf{76.89} }&\scriptsize{76.70} &
 & 
\scriptsize{78.19} & \scriptsize{79.33} & \scriptsize{78.86} & \scriptsize{78.66} & \scriptsize{78.55} & \scriptsize{78.97} & \scriptsize{\textbf{79.64}} & \scriptsize{77.94} & \scriptsize{73.94} & \scriptsize{76.88} &\scriptsize{68.65 }&\scriptsize{79.06}
\\
\hline 
\centering
	$ \vcenter{\includegraphics[scale=0.1]{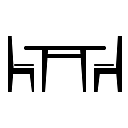}} $ & \scriptsize{73.85} & \scriptsize{75.81} & \scriptsize{74.95} & \scriptsize{\textbf{75.89}} & \scriptsize{75.10} & \scriptsize{75.55} & \scriptsize{75.41} & \scriptsize{73.12} & \scriptsize{68.97} & \scriptsize{74.53} &\scriptsize{75.16 }&\scriptsize{75.07} &
 & 
\scriptsize{76.04} & \scriptsize{76.55} & \scriptsize{75.40} & \scriptsize{75.73} & \scriptsize{75.97} & \scriptsize{76.55} & \scriptsize{76.43} & \scriptsize{73.35} & \scriptsize{68.45} & \scriptsize{76.50} &\scriptsize{71.19 }& \scriptsize{\textbf{76.70}}
\\
\hline 
\centering
	$ \vcenter{\includegraphics[scale=0.1]{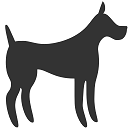}} $ & \scriptsize{85.22} & \scriptsize{\textbf{87.47}} & \scriptsize{86.69} & \scriptsize{86.39} & \scriptsize{86.93} & \scriptsize{86.45} & \scriptsize{86.46} & \scriptsize{86.08} & \scriptsize{84.85} & \scriptsize{86.60} &\scriptsize{ 86.55}& \scriptsize{87.16} &
 & 
\scriptsize{92.06} & \scriptsize{\textbf{92.31}} & \scriptsize{92.31} & \scriptsize{91.79} & \scriptsize{92.14} & \scriptsize{92.14} & \scriptsize{92.27} & \scriptsize{91.59} & \scriptsize{89.40} & \scriptsize{92.28} &\scriptsize{89.28 }& \scriptsize{92.25}
\\
\hline 
\centering
	$ \vcenter{\includegraphics[scale=0.1]{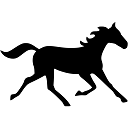}} $ & \scriptsize{87.40} & \scriptsize{87.74} & \scriptsize{87.78} & \scriptsize{87.43} & \scriptsize{87.64} & \scriptsize{87.79} & \scriptsize{87.91} & \scriptsize{86.93} & \scriptsize{86.81} & \scriptsize{87.16} &\scriptsize{87.37 }& \scriptsize{\textbf{87.92}} &
 & 
\scriptsize{88.09} & \scriptsize{\textbf{89.18}} & \scriptsize{88.61} & \scriptsize{88.29} & \scriptsize{89.00} & \scriptsize{87.93} & \scriptsize{88.97} & \scriptsize{87.21} & \scriptsize{86.19} & \scriptsize{87.97} &\scriptsize{ 82.46}& \scriptsize{88.59}
\\
\hline 
\centering
	$ \vcenter{\includegraphics[scale=0.1]{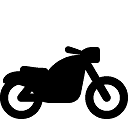}} $ & \scriptsize{85.65} & \scriptsize{85.82} & \scriptsize{85.10} & \scriptsize{84.68} & \scriptsize{85.42} & \scriptsize{85.64} & \scriptsize{85.35} & \scriptsize{84.61} & \scriptsize{84.75} & \scriptsize{85.54} &\scriptsize{ 85.32}& \scriptsize{\textbf{85.87}} &
 & 
\scriptsize{88.71} & \scriptsize{89.29} & \scriptsize{88.89} & \scriptsize{89.07} & \scriptsize{89.24} & \scriptsize{88.86} & \scriptsize{89.28} & \scriptsize{87.89} & \scriptsize{86.69} & \scriptsize{89.38} &\scriptsize{86.69 }& \scriptsize{\textbf{89.59}}
\\
\hline 
\centering
	$ \vcenter{\includegraphics[scale=0.1]{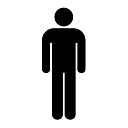}} $ & \scriptsize{92.37} & \scriptsize{92.58} & \scriptsize{91.27} & \scriptsize{92.46} & \scriptsize{92.28} & \scriptsize{92.46} & \scriptsize{\textbf{92.63}} & \scriptsize{92.39} & \scriptsize{89.70} & \scriptsize{92.20} &\scriptsize{ 92.15}& \scriptsize{92.22} &
 & 
\scriptsize{94.24} & \scriptsize{94.37} & \scriptsize{94.04} & \scriptsize{94.21} & \scriptsize{\textbf{94.40}} & \scriptsize{94.31} & \scriptsize{94.37} & \scriptsize{94.02} & \scriptsize{92.75} & \scriptsize{94.24} &\scriptsize{ 91.73}& \scriptsize{93.65}
\\
\hline 
\centering
	$ \vcenter{\includegraphics[scale=0.1]{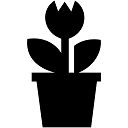}} $ & \scriptsize{58.16} & \scriptsize{\textbf{61.33}} & \scriptsize{57.50} & \scriptsize{58.06} & \scriptsize{58.06} & \scriptsize{58.16} & \scriptsize{60.22} & \scriptsize{56.11} & \scriptsize{50.19} & \scriptsize{60.42} &\scriptsize{57.54 }& \scriptsize{58.13} &
 & 
\scriptsize{55.39} & \scriptsize{\textbf{56.72}} & \scriptsize{54.73} & \scriptsize{55.73} & \scriptsize{56.07} & \scriptsize{55.94} & \scriptsize{56.47} & \scriptsize{52.73} & \scriptsize{43.80} & \scriptsize{55.95} &\scriptsize{46.65 }& \scriptsize{55.48}
\\
\hline 
\centering
	$ \vcenter{\includegraphics[scale=0.1]{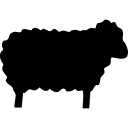}} $ & \scriptsize{81.13} & \scriptsize{81.13} & \scriptsize{80.33} & \scriptsize{82.38} & \scriptsize{83.10} & \scriptsize{80.94} & \scriptsize{80.88} & \scriptsize{77.99} & \scriptsize{79.51} & \scriptsize{79.94} &\scriptsize{ \textbf{83.23}}& \scriptsize{81.88} &
 & 
\scriptsize{81.19} & \scriptsize{82.04} & \scriptsize{80.65} & \scriptsize{80.78} & \scriptsize{81.37} & \scriptsize{81.45} & \scriptsize{\textbf{82.39}} & \scriptsize{79.72} & \scriptsize{78.97} & \scriptsize{81.77} &\scriptsize{76.39 }&\scriptsize{81.37}
\\
\hline 
\centering
	$ \vcenter{\includegraphics[scale=0.1]{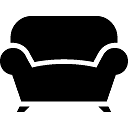}} $ & \scriptsize{67.03} & \scriptsize{67.58} & \scriptsize{65.01} & \scriptsize{67.53} & \scriptsize{68.35} & \scriptsize{69.10} & \scriptsize{68.19} & \scriptsize{64.58} & \scriptsize{61.50} & \scriptsize{68.25} &\scriptsize{69.30 }& \scriptsize{\textbf{70.87}} &
 & 
\scriptsize{64.67} & \scriptsize{67.14} & \scriptsize{65.71} & \scriptsize{66.12} & \scriptsize{66.20} & \scriptsize{66.00} & \scriptsize{67.21} & \scriptsize{63.13} & \scriptsize{55.83} & \scriptsize{64.90} &\scriptsize{ 60.86}& \scriptsize{\textbf{68.11}}
\\
\hline 
\centering
	$ \vcenter{\includegraphics[scale=0.1]{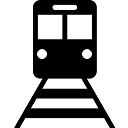}} $ & \scriptsize{92.33} & \scriptsize{91.50} & \scriptsize{92.60} & \scriptsize{92.00} & \scriptsize{92.36} & \scriptsize{92.90} & \scriptsize{92.49} & \scriptsize{91.66} & \scriptsize{91.32} & \scriptsize{\textbf{93.13}} &\scriptsize{92.08 }& \scriptsize{92.50} &
 & 
\scriptsize{94.85} & \scriptsize{94.38} & \scriptsize{\textbf{94.95}} & \scriptsize{94.23} & \scriptsize{94.35} & \scriptsize{94.30} & \scriptsize{94.25} &\scriptsize{93.71} & \scriptsize{94.37} & \scriptsize{94.92} &\scriptsize{ 93.12}& \scriptsize{94.22}
\\
\hline 
\centering
	$ \vcenter{\includegraphics[scale=0.1]{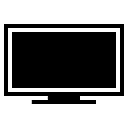}} $ & \scriptsize{76.61} & \scriptsize{76.61} & \scriptsize{76.88} & \scriptsize{76.37} & \scriptsize{76.08} & \scriptsize{77.10} & \scriptsize{76.83} & \scriptsize{74.54} & \scriptsize{73.64} & \scriptsize{77.57} &\scriptsize{ 76.93}& \scriptsize{\textbf{77.62}} &
 & 
\scriptsize{80.63} & \scriptsize{80.43} & \scriptsize{80.67} & \scriptsize{80.56} & \scriptsize{80.24} & \scriptsize{80.77} & \scriptsize{80.57} & \scriptsize{78.83} & \scriptsize{77.41} & \scriptsize{81.54} &\scriptsize{78.24 }& \scriptsize{\textbf{81.80}}
\\
\hline \hline 
\centering
	$ \vcenter{\includegraphics[scale=0.12]{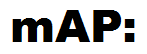}} $ & \scriptsize{80.52} & \scriptsize{\textbf{81.07*}} & \scriptsize{80.25} & \scriptsize{80.59} & \scriptsize{80.90} & \scriptsize{80.83} & \scriptsize{\textbf{80.97*}} & \scriptsize{79.46} & \scriptsize{77.98} & \scriptsize{80.93} &\scriptsize{80.86 }& \scriptsize{\textbf{81.21*}} &
 & 
\scriptsize{82.28} & \scriptsize{\textbf{82.93*}} & \scriptsize{82.42} & \scriptsize{82.48} & \scriptsize{82.61} & \scriptsize{82.65} & \scriptsize{\textbf{82.98*}} & \scriptsize{81.23} & \scriptsize{78.19} & \scriptsize{82.56} &\scriptsize{ 78.86}&\scriptsize{\textbf{82.85*}}
\\
\hline 
\end{tabular}}
\caption{The image classification results achieved in terms of mean average precision (AP) scores while selecting the first $2,048$ (50\%) features. In bold the top score of each class. We indicate with an asterisks the top three methods. }
\label{table:pascalVOC}
\end{table*}

\subsection{Exp. \#1: Deep Representation with pretraining}\label{sec:exp1}\vspace{-0.1cm}
This section proposes a set of tests on the PASCAL VOC-2007~\cite{pascal-voc-2007} and VOC 2012~\cite{pascal-voc-2012} datasets. We want to assess the strengths and weaknesses of using the ILFS in an object recognition classification task. For this reason, we compare our approach against the 11 state-of-the-art FS methods reported in Tab.~\ref{table:compmethods}. This experiment considers as features the cues extracted with a deep convolutional neural networks (CNNs). We selected the pre-trained model called very deep ConvNets~\cite{Simonyan14c}, which performed favorably to the state of the art for classification and detection in the ImageNet Large-Scale Visual Recognition Challenge 2014 (ILSVRC). We use the 4,096-dimension activations of the last layer as image descriptors (i.e., 4,096 features in total). According to the experimental protocol provided by the VOC challenge, a one-vs-rest SVM classifier for each class is trained (where cross-validation is used to find the best parameter C) and evaluated independently. The performance is measured as mean Average Precision (mAP) across all classes. This metric is used rather than the simple classification accuracy because some datasets (particularly the VOC series) were unbalanced in class cardinality. mAP is calculated according to the standard evaluation protocol which involves the use of the \emph{PASCAL VOC Evaluation Server}. As for the Inf-FS, we set its parameters without any cross-validation (i.e., $\alpha=0.2$). Tab.~\ref{table:pascalVOC} serves to analyze how well important features are ranked high by several FS algorithms. The number of features used for both the experiments is set to: $50\%$ of the total. The results are significant: our method achieved the best performance in terms of mean average precision (mAP) on the VOC-2007, followed by Fisher, MI. In the same way, results on VOC-12 shows that the ILFS is still one of the first three best approaches, namely: MI, Fisher, and ours. This set of FS methods achieved the best performance compared with the others, moreover, according to the overall performance over both VOC datasets the methods can be ranked as: \textit{ILFS}, Fisher, and MI. However, it is not possible to infer which one of them performs better to a statistically significant extent (see Sec.\ref{sec:stats} for further details). 
\vspace{-0.2cm}

\subsection{Exp.\#2: Miscellaneous Datasets}\label{sec:exp2}\vspace{-0.2cm}

In this section we provide results obtained on 8 different publicly available benchmarks provided without a particular definition of what the training, validation and testing set are. Therefore, the experimental protocol used in this section consists in splitting the dataset up to 2/3 for training and 1/3 for testing. In order to avoid any biases given for a particular favorable split, this procedure is repeated for 20 times and results are averaged over the trials. Accordingly, each method has been compared against all the others on the same splits for a fair comparison. Feature selection is applied only on the training set and features are selected, generating different subsets of different cardinality (i.e., 10, 50, 100, 150, and 200). As for the previous scenario, the classification is performed using a linear SVM, where a 5-fold cross validation on training data is used to set the best parameters. Results are reported in terms of mAP as for the previous experiment. 
\begin{table*}[t!]
\begin{center}
\resizebox{0.85\textwidth}{!}{%
\begin{tabular}{|l|C{1.1cm}|C{1.1cm}|C{1.1cm}|C{1.3cm}|C{0.9cm}||C{1.7cm}|||C{0.9cm}|C{0.9cm}|C{1.1cm}||C{1.7cm}|}
\hline
 & \multicolumn{6}{c|||}{\textbf{DNA Microarray data}}                                                                                         & \multicolumn{4}{c|}{\textbf{Data from other sources}}                                                                                         \\ \hline
\multicolumn{1}{|c|}{\scriptsize{\textbf{Methods}}}   & \scriptsize{COLON}     & \scriptsize{LEUKEMIA}     & \scriptsize{PROSTATE}     & \scriptsize{LYMPHOMA}     & \scriptsize{LUNG} & \scriptsize{\textbf{Average} [Min,Max]]} &  \scriptsize{GINA}     & \scriptsize{DEXTER}     & \scriptsize{POLARITY}    & \scriptsize{\textbf{Average} [Min,Max]}\\ \hline\hline
\scriptsize{CFS}
&  \scriptsize{81.25 $\qquad\pm$ 0.08} & \scriptsize{96.27 $\qquad\pm$ 0.06} & \scriptsize{85.00 $\qquad\pm$ 0.08 } & \scriptsize{ 84.00  $\qquad\pm$ 0.10}  & \scriptsize{94.50 $\qquad\pm$ 0.17} & \scriptsize{ 88.20  [81.25,96.27]} & 
\scriptsize{81.91 $\qquad\pm$ 0.11} & \scriptsize{79.56 $\qquad\pm$ 0.06}& \scriptsize{86.99 $\qquad\pm$ 0.05} & \scriptsize{ 82.82 [79.56,86.99]}  \\ \hline
\scriptsize{Fisher} 
&  \scriptsize{87.83 $\qquad\pm$ 0.05} & \scriptsize{95.21 $\qquad\pm$ 0.006} & \scriptsize{93.55 $\qquad\pm$ 0.03} & \scriptsize{94.62 $\qquad\pm$ 0.05}  & \scriptsize{97.75 $\qquad\pm$ 0.06} & \scriptsize{ 93.79 [87.83,97.75]} & 
\scriptsize{\textbf{89.36* $\qquad\pm$ 0.03}} & \scriptsize{95.65 $\qquad\pm$ 0.06}& \scriptsize{82.61 $\qquad\pm$ 0.13} & \scriptsize{89.20 [82.61,95.65]}  \\ \hline
\scriptsize{FSV} 
&  \scriptsize{88.00 $\qquad\pm$ 0.05} & \scriptsize{91.57 $\qquad\pm$ 0.01} & \scriptsize{93.50 $\qquad\pm$ 0.02} & \scriptsize{89.38 $\qquad\pm$ 0.04}  & \scriptsize{\textbf{98.83 $\qquad\pm$ 0.01}} & \scriptsize{ 92.25 [88.00,98.83]} & 
\scriptsize{81.73 $\qquad\pm$ 0.12} & \scriptsize{96.39 $\qquad\pm$ 0.01}& \scriptsize{86.12 $\qquad\pm$ 0.12} & \scriptsize{88.08 [81.73,96.39]}  \\ \hline
\scriptsize{LLCFS}
&  \scriptsize{90.00 $\qquad\pm$ 0.05} & \scriptsize{\textbf{99.37 $\qquad\pm$ 0.02}} & \scriptsize{85.80 $\qquad\pm$ 0.09} & \scriptsize{ 84.12 $\qquad\pm$ 0.11}  & \scriptsize{97.69 $\qquad\pm$ 0.04} & \scriptsize{91.39 [84.12,99.37]} & 
\scriptsize{81.91 $\qquad\pm$ 0.09} & \scriptsize{84.16 $\qquad\pm$ 0.10}& \scriptsize{\textbf{97.31 $\qquad\pm$ 0.02}} & \scriptsize{87.79 [81.91,97.31]}  \\ \hline
\scriptsize{LS}
&  \scriptsize{91.58 $\qquad\pm$ 0.05} & \scriptsize{93.57 $\qquad\pm$ 0.006} & \scriptsize{82.00 $\qquad\pm$ 0.12} & \scriptsize{78.88 $\qquad\pm$ 0.16}  & \scriptsize{97.81 $\qquad\pm$ 0.06} & \scriptsize{88.76 [78.88,97.81]} & 
\scriptsize{78.10 $\qquad\pm$ 0.08} & \scriptsize{85.25 $\qquad\pm$ 0.12}& \scriptsize{\textbf{97.77* $\qquad\pm$ 0.02}} & \scriptsize{ 87.04 [78.10,97.77]}  \\ \hline
\scriptsize{MCFS}
&  \scriptsize{90.92 $\qquad\pm$ 0.05} & \scriptsize{92.00 $\qquad\pm$ 0.02} & \scriptsize{76.75 $\qquad\pm$ 0.08} & \scriptsize{84.38 $\qquad\pm$ 0.09}  & \scriptsize{96.53 $\qquad\pm$ 0.16} & \scriptsize{ 88.11 [76.75,96.53]} & 
\scriptsize{85.69 $\qquad\pm$ 0.07} & \scriptsize{87.80 $\qquad\pm$ 0.07}& \scriptsize{95.26 $\qquad\pm$ 0.03} & \scriptsize{89.58 [85.69,95.26]}  \\ \hline
\scriptsize{MI}
&  \scriptsize{86.92 $\qquad\pm$ 0.05} & \scriptsize{93.36 $\qquad\pm$ 0.04} & \scriptsize{90.50 $\qquad\pm$ 0.04} & \scriptsize{94.00 $\qquad\pm$ 0.04}  & \scriptsize{98.72 $\qquad\pm$ 0.02} & \scriptsize{92.70 [86.92,98.72]} & 
\scriptsize{88.85 $\qquad\pm$ 0.04} & \scriptsize{59.51 $\qquad\pm$ 0.04}& \scriptsize{56.19 $\qquad\pm$ 0.09} & \scriptsize{ 68.18 [56.19,88.85]}  \\ \hline
\scriptsize{ReliefF}
&  \scriptsize{84.75 $\qquad\pm$ 0.07} & \scriptsize{93.07 $\qquad\pm$ 0.02} & \scriptsize{93.25 $\qquad\pm$ 0.04} & \scriptsize{91.75 $\qquad\pm$ 0.05}  & \scriptsize{97.33 $\qquad\pm$ 0.03} & \scriptsize{92.03 [84.75,97.33]} & 
\scriptsize{88.86  $\qquad\pm$ 0.03} & \scriptsize{89.54 $\qquad\pm$ 0.12}& \scriptsize{95.82 $\qquad\pm$ 0.03} & \scriptsize{91.40 [88.86,95.82]}  \\ \hline
\scriptsize{RFE }   
&  \scriptsize{82.58 $\qquad\pm$ 0.09} & \scriptsize{86.43 $\qquad\pm$ 0.07} & \scriptsize{78.90 $\qquad\pm$ 0.10} & \scriptsize{77.50 $\qquad\pm$ 0.12}  & \scriptsize{94.25 $\qquad\pm$ 0.17} & \scriptsize{84.53 [77.50,94.25]} & 
\scriptsize{83.05 $\qquad\pm$ 0.09 } & \scriptsize{87.38 $\qquad\pm$ 0.09}& \scriptsize{94.20 $\qquad\pm$ 0.02} & \scriptsize{88.21 [83.05,94.20]}  \\ \hline
\scriptsize{UDFS}   
&  \scriptsize{88.00 $\qquad\pm$ 0.05} & \scriptsize{89.21 $\qquad\pm$ 0.07} & \scriptsize{84.25 $\qquad\pm$ 0.08} & \scriptsize{80.50 $\qquad\pm$ 0.12}  & \scriptsize{ 96.36 $\qquad\pm$ 0.13} & \scriptsize{87.66 [80.50,96.36]} & 
\scriptsize{72.28 $\qquad\pm$ 0.11} & \scriptsize{80.40 $\qquad\pm$ 0.12}& \scriptsize{87.43 $\qquad\pm$ 0.08} & \scriptsize{80.03 [72.28,87.43]}  \\ \hline
\scriptsize{Inf-FS}   
&  \scriptsize{\textbf{96.10 $\qquad\pm$ 0.05}} & \scriptsize{\textbf{99.44} $\qquad\pm$ 0.008} & \scriptsize{92.10 $\qquad\pm$ 0.07} & \scriptsize{96.50 $\qquad\pm$ 0.06}  & \scriptsize{ 97.36 $\qquad\pm$ 0.06} & \scriptsize{96.30 [92.10,99.44]} & 
\scriptsize{78.97 $\qquad\pm$ 0.04} & \scriptsize{81.95 $\qquad\pm$ 0.08}& \scriptsize{68.88 $\qquad\pm$ 0.09} & \scriptsize{76.60 [68.88,81.95]}  \\ \hline
\scriptsize{\textbf{Ours}}
&  \scriptsize{\textbf{96.35* $\qquad\pm$ 0.05}} & \scriptsize{\textbf{99.60* $\qquad\pm$ 0.007}} & \scriptsize{\textbf{97.35* $\qquad\pm$ 0.03}} & \scriptsize{\textbf{99.00* $\qquad\pm$ 0.03}}  & \scriptsize{\textbf{98.98* $\qquad\pm$ 0.03}} & \scriptsize{ \textbf{98.25*} [\textbf{96.35},\textbf{99.60}]} & 
\scriptsize{\textbf{89.03  $\qquad\pm$ 0.03}} & \scriptsize{\textbf{97.81* $\qquad\pm$ 0.01}}& \scriptsize{\textbf{97.76 $\qquad\pm$ 0.01}} & \scriptsize{\textbf{94.87*} [\textbf{89.03},\textbf{97.81}]}  \\ \hline
\end{tabular}}
\caption{Performance of Feature Selection Methods. Average performance obtained with the first 10, 50, 100, 150, and 200 features. The final results are expressed as mean Average Precision (mAP) and their standard deviation. Furthermore, ``\textbf{*}" indicates the top performance.} 
\label{tab_BIO2}
\end{center}
\vspace{-1.5mm}
\end{table*}
Tab.~\ref{tab_BIO2} lists the mAP obtained by averaging the results of the different cardinality. As for the Inf-FS, we set its parameters without any cross-validation (i.e., $\alpha=0.2$). Results show that our approach is very robust across all datasets. All the other methods show a high performance on some datasets and low on others. For example, MI is very close to a random performance on POLARITY and DEXTER, thereby indicating a weakness of the method when applied to sparse data (see Tab.~\ref{table:ch5datasets}). The ILFS is not affected by this problem, and it achieves the best significant performance on DEXTER ($\approx 20K$ features) and a high performance on POLARITY. Fisher, which performs well over all the datasets does not show the same ranking quality as ILFS. Tab.~\ref{tab_BIO2} also reports the overall average scores across the datasets, which clearly show that our approach outperforms all the competitors at all the features' cardinality. Min/Max values are reported in Table~\ref{tab_BIO2} to highlight the robustness of the ILFS to different datasets. In particular, on DNA Microarray data the overall minimum value reported by the ILFS is $+8.35\%$ over the second best (FSV). As for the other datasets, the ILFS still represents the top scoring method according to its overall average, minimum, and maximum scores. 

\vspace{-0.2cm}
\subsection{Reliability and Validity}\label{sec:stats}\vspace{-0.2cm}

In order to assess whether the difference in performance is statistically significant, a set of Student's t-test have been applied to the results~\cite{anderson1984multivariate}. We use the statistical tests to determine if the accuracy given by the proposed approach is significantly different from the one of the other methods (whereas both the distribution of values were normal). The test for assessing whether the data come from normal distributions with unknown, but equal, variances is the \emph{Lilliefors} test \cite{conover1980practical}. Each accuracy reported in Tab.~\ref{tab_BIO2} comes from the average of the accuracies obtained from a series of SVM classifications over 20 different splits of the data for $5$ different subsets of features (i.e., a total of 100 different tests for each method). Thus, given the distribution of these accuracies for the proposed method $d_p$, and the ones of the $i^{th}$ competitor $d_{c_i}$, a \textit{two-sample t-test} has been applied obtaining a test decision for the \emph{null hypothesis} $H_0$ that all the data come from independent random samples from normal distributions. As for the object recognition task (see Tab.~\ref{table:pascalVOC}), we consider as $d_p$ the distribution of accuracies obtained over the 20 classes, and then we compare this distribution against the ones of all the other methods $d_{c_i}$. 
From each t-test we consider the probability (p-value) at which the null hypothesis $H_0$ can be rejected. Based on this result, we assess the validity of the reported results by the \emph{binomial cumulative distribution} function~\cite{anderson1984multivariate,conover1980practical}. We consider $N=10$ independent experiments (i.e., one for each dataset) with exactly two possible outcomes: success and failure. Success when the ILFS outperforms all the other methods with a certain probability to do it by chance $p$. From Tab.~\ref{tab_BIO2} and Tab.~\ref{table:pascalVOC} we observe $k=7$ successes where $p$ is given by the exact p-value at which $H_0$ can be rejected. Since our approach is tested 10 times in the experiments and has $p$ of probability of outperforming the competitors by chance, then the probability of ILFS outperforming more than $k$ times by chance is $4.82 \cdot 10^{-3}$. In conclusion, our approach achieved top performance across many different datasets and difficulties. 

\vspace{-0.3cm}

%
%
\section{Conclusion}\label{sec:conc}\vspace{-0.2cm}

In this paper we proposed a probabilistic feature selection algorithm that performs the ranking step by considering all the possible subsets of features bypassing the combinatorial problem. The most appealing characteristic of the ILFS is that it aims to model the features ``relevancy" using PLSA-inspired process. The derived mixing weights $P(z|f)$ are used to weight a graph of features. The weighted graph, serves to perform the ranking step providing a score of importance for each feature as a function of the importance of its neighbors. Our approach overcomes all the methods in comparison in terms of robustness and ranking quality in a statistically significant extent, attaining the highest performance levels across all the challenging scenarios and difficulties. This study also points to many future directions. From a methodological perspective, the investigation of the absorbing Markov chains has every opportunity to reveal a criterion to perform the \emph{subset selection} step automatically. 
Results of our work can possibly be improved by performing a validation over multiple $\mathcal{T}$ intervals. As for the applications, we hope that this work motivates researchers to take into account the use of FS as an integral part of future computer vision systems. Finally, for the sake of repeatability, the source code is available at \url{https://goo.gl/uTuZhc} to provide the material needed to replicate our experiments.

\section*{Acknowledgements}\vspace{-0.3cm}
\noindent
This work is supported in part by EPSRC under grants EP/N035305/1 and EP/M025055/1.

{\small
\bibliographystyle{ieee}
\bibliography{egbib}

\begin{thebibliography}{10}\itemsep=-1pt

\bibitem{citeulike:1624492}
{Gene expression correlates of clinical prostate cancer behavior}.
\newblock {\em Cancer Cell}, 1(2):203--209, 2002.

\bibitem{GINA}
{GINA} digit recognition database {IJCNN}.
\newblock 2007.

\bibitem{anderson1984multivariate}
T.~Anderson.
\newblock Multivariate statistical analysis.
\newblock {\em VVi11ey and Sons, New York, NY}, 1984.

\bibitem{Bergshoeff}
E.~Bergshoeff.
\newblock Ten physical applications of spectral zeta functions.
\newblock {\em CQG}, 13(7), 1996.

\bibitem{FS_ICCV1}
J.~Bins and B.~A. Draper.
\newblock Feature selection from huge feature sets.
\newblock In {\em Conf. IEEE International Conference on Computer Vision},
  volume~2, pages 159--165 vol.2, 2001.

\bibitem{Borgatti2005}
S.~P. Borgatti and M.~G. Everett.
\newblock {A Graph-theoretic perspective on centrality}.
\newblock {\em Social Networks}, 28(4):466--484, 2006.

\bibitem{Bradley98featureselection}
P.~S. Bradley and O.~L. Mangasarian.
\newblock Feature selection via concave minimization and support vector
  machines.
\newblock In {\em ICML}, pages 82--90. Morgan Kaufmann, 1998.

\bibitem{Cai:2010}
D.~Cai, C.~Zhang, and X.~He.
\newblock Unsupervised feature selection for multi-cluster data.
\newblock In {\em Proceedings of the 16th ACM SIGKDD International Conference
  on Knowledge Discovery and Data Mining}, pages 333--342, 2010.

\bibitem{conover1980practical}
W.~J. Conover and W.~J. Conover.
\newblock Practical nonparametric statistics.
\newblock 1980.

\bibitem{FS_ICCV3}
G.~Dorko and C.~Schmid.
\newblock Selection of scale-invariant parts for object class recognition.
\newblock In {\em Conf. IEEE International Conference on Computer Vision},
  pages 634--639, 2003.

\bibitem{pascal-voc-2007}
M.~Everingham, L.~Van~Gool, C.~K.~I. Williams, J.~Winn, and A.~Zisserman.
\newblock The {PASCAL} {V}isual {O}bject {C}lasses {C}hallenge 2007 {(VOC2007)}
  {R}esults.

\bibitem{pascal-voc-2012}
M.~Everingham, L.~Van~Gool, C.~K.~I. Williams, J.~Winn, and A.~Zisserman.
\newblock The {PASCAL} {V}isual {O}bject {C}lasses {C}hallenge 2012 {(VOC2012)}
  {R}esults.

\bibitem{FS_ICCV2}
Z.-G. Fan and B.-L. Lu.
\newblock Fast recognition of multi-view faces with feature selection.
\newblock In {\em Conf. IEEE International Conference on Computer Vision},
  volume~1, pages 76--81, 2005.

\bibitem{Golub99}
T.~R. e.~a. Golub.
\newblock Molecular classification of cancer: class discovery and class
  prediction by gene expression monitoring.
\newblock {\em Science}, 286:531--537, 1999.

\bibitem{Gordon02}
G.~J. Gordon, R.~V. Jensen, L.~li~Hsiao, S.~R. Gullans, J.~E. Blumenstock,
  S.~Ramaswamy, W.~G. Richards, D.~J. Sugarbaker, and R.~Bueno.
\newblock Translation of microarray data into clinically relevant cancer
  diagnostic tests using gene expression ratios in lung cancer and
  mesothelioma.
\newblock {\em Cancer Res}, 62:4963--4967, 2002.

\bibitem{Graham:1994}
R.~L. Graham, D.~E. Knuth, and O.~Patashnik.
\newblock {\em Concrete Mathematics: A Foundation for Computer Science}.
\newblock Addison-Wesley, 1994.

\bibitem{Quanquanjournals}
Q.~Gu, Z.~Li, and J.~Han.
\newblock Generalized fisher score for feature selection.
\newblock {\em CoRR}, abs/1202.3725, 2012.

\bibitem{NIPS2003}
I.~Guyon, J.~Li, T.~Mader, P.~A. Pletscher, G.~S. 0004, and M.~Uhr.
\newblock Competitive baseline methods set new standards for the {NIPS} 2003
  feature selection benchmark.
\newblock {\em PRL}, 28(12):1438--1444, 2007.

\bibitem{Guyon:2002}
I.~Guyon, J.~Weston, S.~Barnhill, and V.~Vapnik.
\newblock Gene selection for cancer classification using support vector
  machines.
\newblock {\em Mach. Learn.}, 46(1-3):389--422, 2002.

\bibitem{HCN05a}
X.~He, D.~Cai, and P.~Niyogi.
\newblock Laplacian score for feature selection.
\newblock In {\em Advances in Neural Information Processing Systems 18}, 2005.

\bibitem{hofmann1999probabilistic}
T.~Hofmann.
\newblock Probabilistic latent semantic analysis.
\newblock In {\em Proceedings of the Fifteenth conference on Uncertainty in
  artificial intelligence}, pages 289--296. Morgan Kaufmann Publishers Inc.,
  1999.

\bibitem{HubHub01}
J.~H. Hubbard and B.~B. Hubbard, editors.
\newblock {\em Vector Calculus, Linear Algebra, and Differential Forms: A
  Unified Approach (Edition 2)}.
\newblock Pearson, 2001.

\bibitem{liu2008}
H.~Liu and H.~Motoda.
\newblock {\em Computational Methods of Feature Selection}.
\newblock Chapman and Hall, 2008.

\bibitem{ICCV_4}
X.~Liu and T.~Yu.
\newblock Gradient feature selection for online boosting.
\newblock In {\em Conf. IEEE International Conference on Computer Vision},
  pages 1--8, 2007.

\bibitem{KristanLMFPCVHL16}
K.~Matej and et~Al.
\newblock The visual object tracking {VOT2016} challenge results.
\newblock In {\em Conf. IEEE European Conference on Computer Vision,
  Workshops}, pages 777--823, 2016.

\bibitem{ReviewPangLee2004}
B.~Pang and L.~Lee.
\newblock A sentimental education: Sentiment analysis using subjectivity
  summarization based on minimum cuts.
\newblock In {\em Proceedings of the ACL}, 2004.

\bibitem{roffo2016feature}
G.~Roffo.
\newblock Feature selection library (matlab toolbox).
\newblock {\em arXiv preprint arXiv:1607.01327}, 2016.

\bibitem{RoffoBMVC2016}
G.~Roffo and S.~Melzi.
\newblock Online feature selection for visual tracking.
\newblock In {\em Conf. The British Machine Vision Conference (BMVC)},
  September 2016.

\bibitem{Roffo2017b}
G.~Roffo and S.~Melzi.
\newblock {\em Ranking to Learn}, pages 19--35.
\newblock Springer International Publishing, Cham, 2017.

\bibitem{Roffo:InfFS:2015}
G.~Roffo, S.~Melzi, and M.~Cristani.
\newblock Infinite feature selection.
\newblock In {\em Conf. IEEE International Conference on Computer Vision},
  pages 4202--4210, 2015.

\bibitem{Simonyan14c}
K.~Simonyan and A.~Zisserman.
\newblock Very deep convolutional networks for large-scale image recognition.
\newblock {\em CoRR}, abs/1409.1556, 2014.

\bibitem{alon}
{U., Alon et Al}.
\newblock Broad patterns of gene expression revealed by clustering analysis of
  tumor and normal colon tissues probed by oligonucleotide arrays.
\newblock In {\em PNAS}, volume~96. 1999.

\bibitem{matrixInversion2}
K.~Wu, C.~Soci, P.~P. Shum, and N.~I. Zheludev.
\newblock Computing matrix inversion with optical networks.
\newblock {\em Opt. Express}, 22(1):295--304, 2014.

\bibitem{Yang:2011}
Y.~Yang, H.~T. Shen, Z.~Ma, and et~Al.
\newblock L2,1-norm regularized discriminative feature selection for
  unsupervised learning.
\newblock In {\em Conf. International Joint Conference on Artificial
  Intelligence}, pages 1589--1594, 2011.

\bibitem{Hutter:02feature}
M.~Zaffalon and M.~Hutter.
\newblock Robust feature selection using distributions of mutual information.
\newblock In {\em UAI}, pages 577--584, 2002.

\bibitem{zeng2011feature}
H.~Zeng and Y.-m. Cheung.
\newblock Feature selection and kernel learning for local learning-based
  clustering.
\newblock {\em IEEE Transactions on Pattern Analysis and Machine Intelligence},
  33(8):1532--1547, 2011.

\end{thebibliography}
}

\end{document}